\documentclass[journal]{IEEEtran}
\ifCLASSINFOpdf
\else
\fi
\usepackage{graphicx}
\usepackage{epsfig}
\usepackage{epstopdf}
\usepackage{float}
\usepackage{algorithmic}
\usepackage{algorithm}
\usepackage{cite}
\usepackage{amsmath,amssymb,amsfonts}
\usepackage{algorithmic}
\usepackage{algorithm}
\usepackage{epsfig}
\usepackage{epstopdf}
\usepackage{graphicx}
\usepackage{textcomp}
\usepackage{xcolor}
\usepackage{lipsum}
\usepackage{stfloats}


\begin{document}
%
\title{A Discriminative Vectorial Framework for Multi-modal Feature Representation}

%

\author{Lei~Gao,~\IEEEmembership{Member,~IEEE,}
        and~Ling~Guan,~\IEEEmembership{Fellow,~IEEE}
\thanks{L. Gao and L. Guan are with the Department of Electrical and Computer Engineering, Ryerson University, Toronto, ON M5B 2K3, Canada (email:iegaolei@gmail.com; lguan@ee.ryerson.ca).}}

\maketitle

\begin{abstract}
Due to the rapid advancements of sensory and computing technology, multi-modal data sources that represent the same pattern or phenomenon have attracted growing attention. As a result, finding means to explore useful information from these multi-modal data sources has quickly become a necessity. In this paper, a discriminative vectorial framework is proposed for multi-modal feature representation in knowledge discovery by employing multi-modal hashing (MH) and discriminative correlation maximization (DCM) analysis. Specifically, the proposed framework is capable of minimizing the semantic similarity among different modalities by MH and exacting intrinsic discriminative representations across multiple data sources by DCM analysis jointly, enabling a novel vectorial framework of multi-modal feature representation. Moreover, the proposed feature representation strategy is analyzed and further optimized based on canonical and non-canonical cases, respectively. Consequently, the generated feature representation leads to effective utilization of the input data sources of high quality, producing improved, sometimes quite impressive, results in various applications. The effectiveness and generality of the proposed framework are demonstrated by utilizing classical features and deep neural network (DNN) based features with applications to image and multimedia analysis and recognition tasks, including data visualization, face recognition, object recognition; cross-modal (text-image) recognition and audio emotion recognition. Experimental results show that the proposed solutions are superior to state-of-the-art statistical machine learning (SML) and DNN algorithms.
\end{abstract}

\begin{IEEEkeywords}
knowledge discovery, multi-modal feature representation, multi-modal hashing, discriminative correlation maximization, image analysis and recognition, cross-modal analysis, audio emotion recognition.
\end{IEEEkeywords}

%
\IEEEpeerreviewmaketitle

\section{Introduction}

\IEEEPARstart{K}{nowledge} discovery is a process of exploring useful information in large volumes of data. It plays a key role in image analysis and synthesis, optimization, high-performance computing and many other tasks [1-4, 97-98]. Nowadays, as sensory and computing technology has rapidly developed, the same pattern or phenomenon can be described from different types of acquisition techniques or even heterogeneous sensors. As a result, multi-modal information analysis has grown at an extremely rapid pace [5-6, 114-115], clearly demonstrating its potential to improve system performance by utilizing the content across multiple data sources.\\\indent Nevertheless, as benefits often come at a cost, new challenges in multi-modal information/data analysis also await us. As a central component in knowledge discovery, feature representation transforms the original features into more effective representations with improved performance [4]. In essence, feature representation aims to explore appropriate representations embedded in the data sets for different applications. In recent years, deep neural networks (DNNs) have shown remarkable performance in image computation, classification and other visual information processing tasks [7-8], particularly for feature representation learning in knowledge discovery [74]. Therefore, the development on feature representation has drawn enormous attention in both academia and industry sectors.\\\indent As one of the important descriptors in multi-modal information analysis, multi-modal hashing (MH) has attracted broad interest and the criterion of semantic similarity is widely applied to multimedia retrieval on data sets of different scales, leading to many successes [113]. Generally, the element-wise `sign' function is utilized for evaluation of the semantic similarity. Since outputs of the `sign' function are two different directions (+1 or -1), essentially, semantic similarity provides a direction criterion to guide the accomplishment of multimedia retrieval among various data sources [88]. While the existing algorithms have generated promising performance, only the semantic similarity amongst modalities is employed; without considering the intrinsic discriminative representations across modalities. It leads to an even bigger challenge: jointly exploring semantic similarity and intrinsic discriminative representations from different modalities in information fusion.\\\indent In this paper, a discriminative vectorial framework is proposed for multi-modal feature representation. By this framework, both multi-modal hashing (MH) and discriminative correlation maximization (DCM) analysis are exploited to generate a discriminative vectorial vehicle of multi-modal feature representation, providing an effective solution to the aforementioned challenges. Not only is the minimum of the semantic similarity among modalities achieved by MH learning, but the intrinsic discriminative representation is also explored by maximizing the within-class correlation and minimizing the between-class correlation, leading to a high-quality feature representation. Thus, the generated feature representation results in better utilization of the input multi-modal data sources, achieving improved performance. Frameworks of the existing multi-modal hashing method and the proposed strategy are depicted graphically in Figure. 1.\\\indent
\begin{figure*}[t]
\centering
\includegraphics[height=2.0in,width=6.8in]{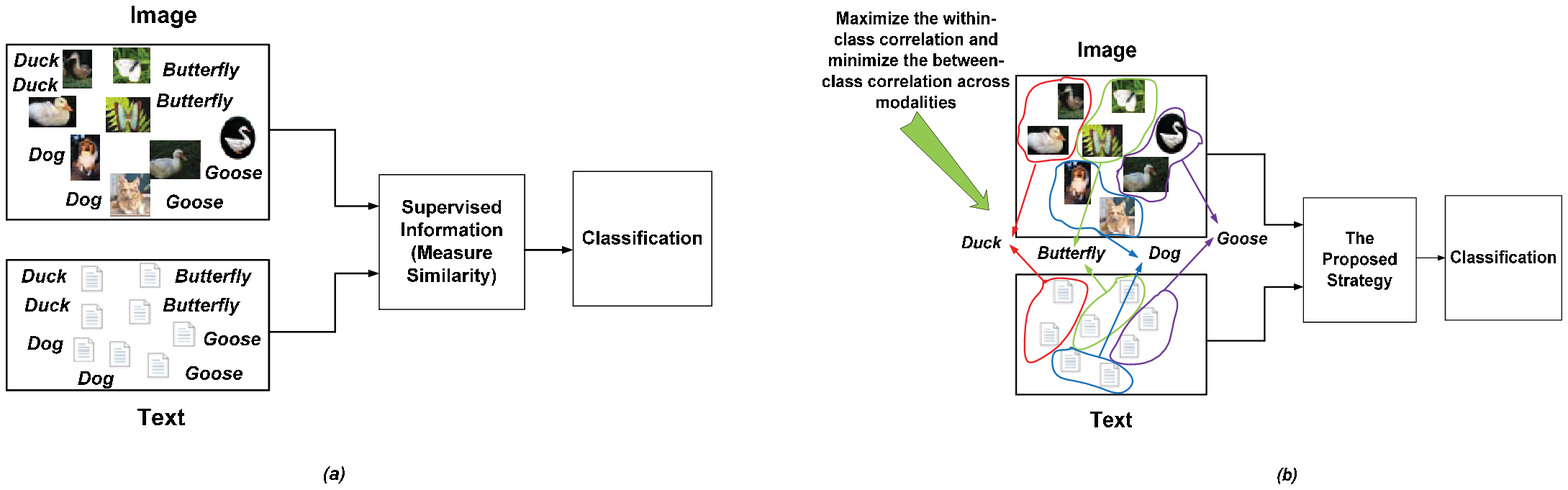}\\ Figure. 1 (a) The framework of the existing multi-modal hashing. (b) The framework of the proposed strategy.\\
\end{figure*}
More importantly, the proposed framework is analyzed and further optimized according to canonical and non-canonical cases respectively, leading to a new way of guiding discriminative projection selection. The generated feature representation is evaluated by comparison with state-of-the-art statistical
machine learning (SML) and DNN algorithms. Note, due to the generic nature of the feature representation framework, the proposed methods are able to handle input features generated by both the classic features and DNN based features. The effectiveness of the proposed framework is demonstrated through applications to data visualization, face recognition, object recognition; cross-modal (text-image) recognition and audio emotion recognition.\\\indent In the following, a review of related work is given in Section II. The proposed discriminative vectorial framework is formulated in Section III. In Section IV, feature extraction is introduced. Experimental results and analysis are presented in Section V. Conclusions are summarized in Section VI.
\section{Related Work}
Currently, as a powerful analysis tool in knowledge discovery, hashing analysis has attracted significant interest in academic and industrial sectors [9, 89-91]. The purpose of hashing is to transform the original input data to a Hamming space with binary hash codes, where the data dimensionality is largely reduced and the query speed can be dramatically improved [10]. Benefiting from low dimensionality and fast query speed, hashing has been applied to visual analysis and recognition, image processing and many other tasks [9-10].\\\indent Hence, the main target of hashing is to accomplish the task of similarity preservation between original input data and the transformed data in a hashing space [11]. To represent the similarity between original and hash space better, the semantic information is adopted widely in hashing analysis with application to various tasks, such as visual classification, large scale data search, etc [12-15]. Lu et al. [12] proposed a latent semantic minimal hashing method to produce the appropriate semantic-preserving binary codes according to query images. Zhu et al. [13] introduced a semantic-assisted visual hashing method to exploit the semantics between auxiliary texts and images for retrieval. Zhou et al. [14] presented a kernel-based semantic hashing model with application in gait retrieval, which can handle the inherent shortcomings of face-based and appearance-based methods. A new discrete semantic transfer hashing algorithm was investigated for image retrieval in [15]. Based on this algorithm, we can extract rich auxiliary contextual information to improve the semantics of discrete codes.\\\indent Although there are extensive investigations on hashing analysis, most of them merely focus on a single-view data set, thereby limiting the applications of this technique. Recently, with the development of multi-modal information analysis, greater effort has been committed to multi-modal hashing. Zheng et al. [16] presented a fast discrete collaborative multi-modal hashing method with applications in large scale multimedia retrieval. Hu et al. [17] studied a collective reconstructive embeddings strategy for multi-modal hashing. Wei et al. [18] presented a heterogeneous translated hashing strategy which can enhance the multi-view search speed as well as improve similarity accuracies within heterogeneous data. Recently, a deep cross-modal hashing model was introduced to multimedia retrieval applications [19]. As an end-to-end framework, it can accomplish feature extraction and hashing transform simultaneously.\\\indent In general, multi-modal hashing is divided into two types: multi-source hashing and cross-modal hashing [20]. The goal of multi-source hashing is to study a better transform strategy by utilizing multi-view data sets rather than a single-view data set [21-22]. Comparatively, cross-modal hashing attracts much more attentions and has found wider applications since only one view is required for a query point [23]. Furthermore, according to the information used in learning, multi-modal hashing is classified into unsupervised multi-modal hashing and supervised multi-modal hashing. As extra supervised information is introduced to supervised multi-modal hashing, it usually achieves better performance than unsupervised techniques. Although most of the existing supervised hashing methods have achieved promising performance in many applications, only the semantic similarity from multiple data sources is utilized, without exploiting intrinsic discriminative representations across different modalities. Therefore, state-of-the-art is still far from satisfactory.\\\indent On the other hand, in the recent past, the correlation analysis based methods have drawn significant attention towards multi-modal information analysis. As a typical representation of correlation analysis-based methods, canonical correlation analysis (CCA) has been applied to multi-modal information fusion, image classification and more [24]. Nevertheless, as an unsupervised method, CCA neither represents the similarity between the samples in the same class, nor effectively evaluates the dissimilarity between the samples in different classes. To tackle the aforementioned problem, the within-class correlation and between-class correlation are utilized jointly to extract more discriminative representations [25]. However, there exists an challenge as how to explore semantic similarity across modalities to gain better generalization ability for feature representation on the small scale data sets.
\section{The Proposed Methods}
In this section, the outline of multi-modal hashing and discriminative correlation maximization analysis is briefly presented, and then used to formulate the proposed methods.
\subsection{The Multi-modal Hashing Method}
Suppose we have two sets of variables ${x} = {[{x_1},...,{x_N}]^T} \in {\rm{ }}{R^{N \times m}}$ and ${y}= {[{y_1},...,{y_N}]^T} \in {R^{\:N \times p}}$ as the inputs, where $m$ and $p$ equal the number of dimensions of $x$ and $y$, and $N$ is the number of training samples. In accordance to the aforementioned categories of multi-modal hashing, supervised multi-modal hashing is the focus of study in this paper.\\\indent It is known that semantic labels for training samples in supervised multi-modal hashing are accessible. Semantic labels are expressed in the form of vectors $\{ {l_1},{l_2}...,{l_N}\left| {{l_i} \in {{\{ 0,1\} }^c}} \right.\} $, where $c$ is the number of all classes. Then, $l_{i,k}=1$ denotes the $i$th sample being in the $k$th class. Otherwise, $l_{i,k}=0$. Suppose we use a matrix $U \in {\rm{ }}{R^{\:N \times c}}$ to consist of semantic labels of training samples with ${U_{i,k}} = {l_{i,k}}$, where $U_{i,k}$ denotes an  element in the $i$th row and $k$th column. Then, the matrix of semantic similarity for multi-modal hashing is formulated in equation (1)
\begin{equation}
S = U{U^T}.
\end{equation}
The purpose of multi-modal hashing is to learn two hashing functions $f()$ and $g()$ across two data sources $f({x_i}):{R^m} \to {\{  - 1,1\} ^L}$ and $g({y_i}):{R^p} \to {\{  - 1,1\} ^L}$ with $L$ being the length of the binary hash codes. While many functions are able to be utilized to define $f(x)$ and $g(y)$, the functions in equation (2) are adopted as
\begin{equation}
\begin{array}{l}
 f(x) = {\mathop{\rm sgn}} (x), \\
 g(y) = {\mathop{\rm sgn}} (y), \\
 \end{array}
\end{equation}
where $sgn()$ is the element-wise sign function. \\\indent 
Since the $sgn$ function owns the following property
\begin{equation}
{\mathop{\rm sgn}}  \in \{  - 1,1\},
\end{equation}
it yields the following relation
\begin{equation}
{\mathop{\rm sgn}}  \cdot {\mathop{\rm sgn}}  \in [ - 1,1].
\end{equation}
A linear element-wise operation is performed on $S$ to obtain the relation in equation (5)
\begin{equation}
S' = 2S - {\bf{1}} \cdot {{\bf{1}}^T},
\end{equation}
where $S' \in {R^{N \times N}}$ and ${\bf{1}} = {[1,1, \cdots 1]^{\bf{T}}} \in {{{R}}^{{N \times 1}}}$.\\\indent
Thus, the solution to the multi-modal hashing is to seek ${W_{x}}$ and ${W_{y}}$ with the minimum semantic similarity among modalities, which is formulated as follows [16]:
\begin{equation}
\mathop {\min }\limits_{{W_x},{W_y}} {\left\| {{\mathop{\rm sgn}} (x{W_x}){\mathop{\rm sgn}} {{(y{W_y})}^T} - L{S^{'}}} \right\|^2}.
\end{equation}
where $||.||$ denotes the norm operation.\\\indent In equation (6), since the `sign' function is utilized for multi-modal hashing, multi-modal feature representation is achieved by a direction (\{--1, 1\}) criterion.
\subsection{The Discriminative Correlation Maximization}
Let ${x} = {[{x_1},...,{x_N}]^T} \in {\rm{ }}{R^{N \times m}}$ and ${y}= {[{y_1},...,{y_N}]^T} \in {R^{\:N \times p}}$ be the two variables sets. The mean vector value of $x$ and $y$ is given in equation (7)\\
\begin{equation}
{x_M} = \frac{1}{N}\sum\limits_{i = 1}^N {{x_i}} ,{y_M} = \frac{1}{N}\sum\limits_{i = 1}^N {{y_i}}.
\end{equation}
Therefore, ${x'}={[{x_1}-x_M,...,{x_N}-x_M]^T}$ and ${y'}= {[{y_1}-y_M,...,{y_N}-y_M]^T}$ are two sets of zero-mean variables, which satisfy the relation in equations (8) and (9)
\begin{equation}
 x'^T \cdot {\bf{1}} = {\bf{0}} \in {R^m},
\end{equation}
\begin{equation}
 y'^T \cdot {\bf{1}} = {\bf{0}} \in {R^p}.
\end{equation}
Suppose ${x'_{i}}^{(d)}$ and ${y'_{i}}^{(d)}$ are the \emph{i}th sample in the \emph{d}th class. ${n_{d}}$ is the number of samples in the \emph{d}th class and it owns the relation in (10)
\begin{equation} \sum\limits_{d = 1}^c {{n_{d}}}  = N. \end{equation}
Then, the within-class correlation is ${C_{{w_{{x'}{y'}}}}={x'}^{T}A{y'}}$ and between-class correlation is ${C_{{b_{{x'}{y'}}}}=- {x'}^TA{y'}}$ [25],
where
\begin{small}
\begin{equation} \ A = \left[ {\left( {\begin{array}{*{20}{c}}{{H_{{n_{1}} \times {n_{1}}}}}& \ldots &0\\
 \vdots &{{H_{{n_{d}} \times {n_{d}}}}}& \vdots \\
0& \ldots &{{H_{_{{n_{c}} \times {n_{c}}}}}}
\end{array}} \right)} \right] \in {R^{N \times N}}, \end{equation} \end{small}
where ${H_{{n_{{d}}} \times {n_{d}}}}$ is in the form of ${n_{d}} \times {n_{d}}$ and all the elements in ${H_{{n_{d}} \times {n_{d}}}}$ are unit values. Therefore, the discriminative multi-modal correlation function is written as follows
\begin{equation}
\mathop {{C_{{x'}{y'}}}}\limits^ \sim   = {C_{{w_{{x'}{y'}}}}} - {C_{{b_{{x'}{y'}}}}}.
\end{equation}
Substituting ${C_{{w_{{x'}{y'}}}}}$ and ${C_{{b_{{x'}{y'}}}}}$ into (12) leads to the following relation
\begin{equation}
\mathop {{C_{x'y'}}}\limits^\sim  = {C_{{w_{x'y'}}}} - {C_{{b_{x'y'}}}} = {x'^T}Ay' - ( - {x'^T}Ay') = 2{x'^T}Ay'.
\end{equation}
In equation (13), it is observed that matrix $A$ plays a central role in extracting discriminative representations. Then, the discriminative correlation maximization is achieved by finding two projected matrices ${W_{x'}}$ and ${W_{y'}}$ in (14)
\begin{equation}
\mathop {\arg \max }\limits_{{W_{x'}},{W_{y'}}} {W_{x'}}^T\mathop {{C_{{x'}{y'}}}}\limits^\sim {W_{y'}}.
\end{equation}
\begin{figure*}[htbp]
\centering
\includegraphics[height=1.5in,width=6.5in]{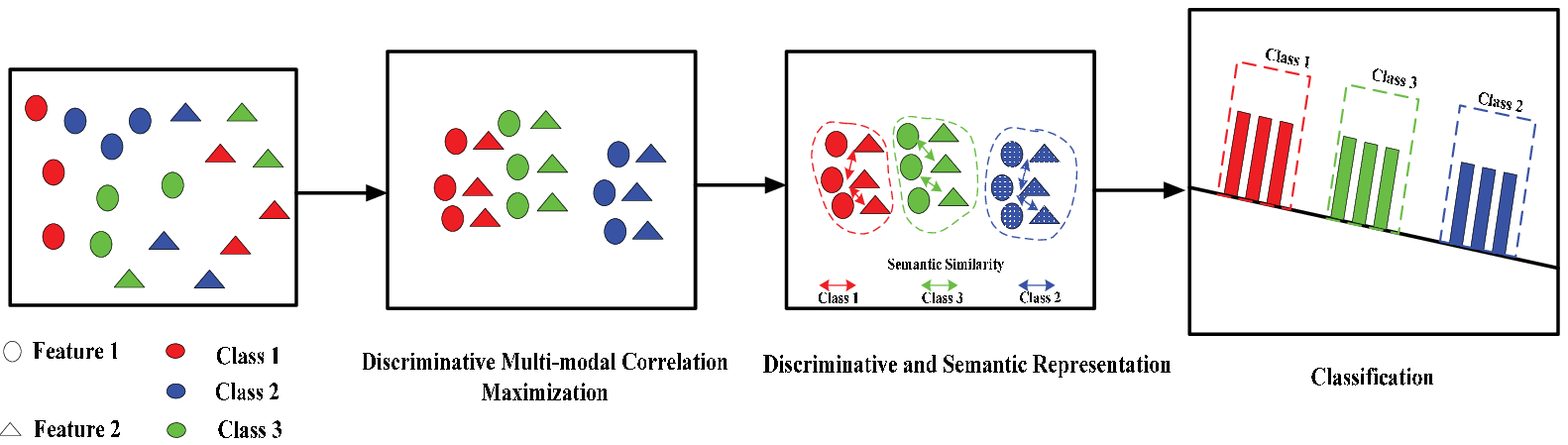}\\ Figure. 3 The representation of the proposed solution.
\end{figure*}
\subsection{Discriminative Vectorial Multi-modal Feature Representation}
From the above subsections, the semantic similarity amongst different modalities is capable of providing a `direction' criterion for effective multi-modal feature representation, but unable to extract discriminative representations effectively.\\\indent On the other side, although the discriminative correlation maximization is capable of exploring more discriminative representations, it only adopts the criterion of `distance' (ie. within-class correlation and between-class correlation in equation (12)), ignoring the direction (semantic similarity) information across different data sources. As a result, it leads to unsatisfied performance, as depicted in Figure 2. In Figure 2, due to a smaller `distance', one sample of class 1 is misclassified into class 3 while another sample from class 3 is misclassified into class 1.
\begin{figure}[H]
\centering
\includegraphics[height=1.1in,width=3.5in]{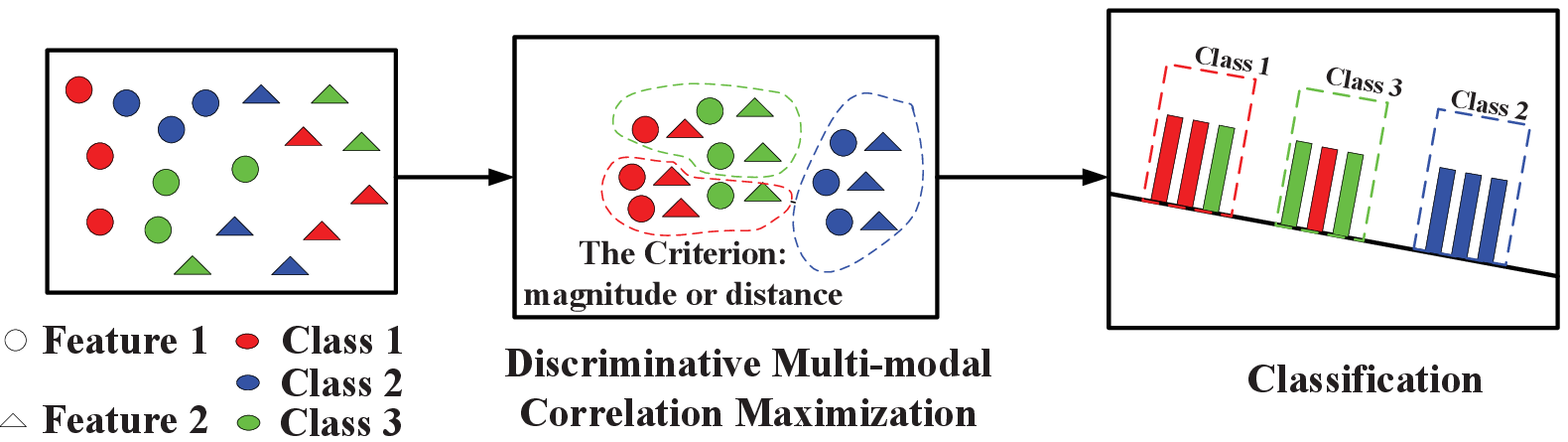}\\ Figure. 2 The representation by discriminative multi-modal correlation maximization
\end{figure}
To address this problem properly, a discriminative vectorial multi-modal feature representation is formulated by integrating the `direction' and `distance'. As the matrix $A$ plays an important role in discriminative representations extraction and satisfies the symmetric property, equation (14) is further written as below
\begin{equation}
\begin{array}{l}
 \mathop {\arg \max }\limits_{{W_{x'}},{W_{y'}}} {W_{x'}}^T\mathop {{C_{{x'}{y'}}}}\limits^\sim {W_{y'}} \\
  = \mathop {\arg \max }\limits_{{W_{x'}},{W_{y'}}} 2{W_{x'}}^T{{x'}^T}A{y'}{W_{y'}} \\
  = \mathop {\arg \max }\limits_{{W_{x'}},{W_{y'}}} 2{({x'}{W_{x'}})^T}A({y'}{W_{y'}}) \\
  = \mathop {\arg \max }\limits_{{W_{x'}},{W_{y'}}} 2{(A^{1/2}{x'}{W_{x'}})^T}(A^{1/2}{y'}{W_{y'}}). \\
 \end{array}
\end{equation}
To integrate the multi-modal hashing and discriminative multi-modal correlation maximization effectively, in this paper, two mapping functions $f()$ and $g()$ are defined as $f(x') = {\mathop{\rm sgn}} (A^{1/2}x'W_{x'})$ and  $g(y') = {\mathop{\rm sgn}} (A^{1/2}y'W_{y'})$. The objective function of the proposed strategy is written as follows
\begin{equation}
\mathop {\min }\limits_{{W_{x'}},{W_{y'}}} {\left\| {{\mathop{\rm sgn}} ({A^{1/2}}x'{W_{x'}}){\mathop{\rm sgn}} {{({A^{1/2}}y'{W_{y'}})}^T} - L{S^{'}}} \right\|^2}.
\end{equation}
Thus, the discriminative and semantic representations among different modalities are optimized jointly to enable a novel discriminative vectorial representation of multi-modal features with both `distance' (discriminative) and `direction' (semantic) components, and thus parallel to the definition of `vector' in Physics. This vectorial representation is illustrated in Figure. 3. \\\indent In the following, the proposed strategy is analyzed and further optimized under the canonical and non-canonical cases, respectively.
\subsubsection{Canonical Correlation Maximization}
Under the canonical condition, two variables sets $x'$ and $y'$ satisfy equation (17) [76]:
\begin{equation}
\begin{array}{l}
 {W_{x'}}^T{x'^T}Ax'{W_{x'}} = N{I_L}, \\
 {W_{y'}}^T{y'^T}Ay'{W_{y'}} = N{I_L}, \\
 \end{array}
\end{equation}
where $I_L$ is an identity matrix and its size is in the form of $L \times L$.
Then, according to the spectral relaxation algorithms in [26], equation (16) is expressed in (18)
\begin{equation}
\mathop {\min }\limits_{{W_x'},{W_y'}} {\left\| {{\mathop{\rm}} ({A^{1/2}}x'{W_x'}){\mathop{\rm}} {{({A^{1/2}}y'{W_y'})}^T} - L{S^{'}}} \right\|^2},
\end{equation}
\quad \quad \quad \quad \emph{s.t.}
\begin{equation}
\begin{array}{l}
 {({A^{1/2}}x'{W_x'})^T}{A^{1/2}}x'{W_x'} = N{I_L}, \\
 {({A^{1/2}}y'{W_y'})^T}{A^{1/2}}y'{W_y'} = N{I_L}. \\
 \end{array}
\end{equation}
Then, equation (18) is further formulated in (20)
\begin{equation}
\begin{array}{l}
 \mathop {\min }\limits_{{W_{x'}},{W_{y'}}} {\left\| {({A^{1/2}}x'{W_{x'}}){{({A^{1/2}}y'{W_{y'}})}^T} - L{S^{'}}} \right\|^2} \\
  =\mathop {\min }\limits_{{W_{x'}},{W_{y'}}} (- Ltr\{ 2[({W_{x'}}^T{x'^T}{A^{1/2}}){S^{'}}({A^{1/2}}y'{W_{y'}})]\} + const.  \\
 \end{array}
\end{equation}
The derivation of equation (20) is given in (20.a).\\
\begin{figure*}[ht]
\hrulefill
\begin{align*} \label{equ:BigWrited LiZi}
\begin{array}{l}
\mathop {\min }\limits_{{W_{x'}},{W_{y'}}} {\left\| {({A^{1/2}}x'{W_{x'}}){{({A^{1/2}}y'{W_{y'}})}^T} - L{S^{'}}} \right\|^2} \\
  = \mathop {\min }\limits_{{W_{x'}},{W_{y'}}} (tr\{ [({A^{1/2}}x'{W_{x'}}){({A^{1/2}}y'{W_{y'}})^T} - L{S^{'}}] {[({A^{1/2}}x'{W_{x'}}){({A^{1/2}}y'{W_{y'}})^T} - L{S^{'}}]^T}\})  \\
  = \mathop {\min }\limits_{{W_{x'}},{W_{y'}}} (- Ltr\{ 2[({W_{x'}}^T{x'^T}{A^{1/2}}){S^{'}}({A^{1/2}}y'{W_{y'}})]\}  + L{N^2} + {L^2}tr\{ [{({S^{'}})^T}{S^{'}}]\}),  \\
 \end{array}
\tag{20.a}
\end{align*}
where $tr()$ is the trace operation of a given matrix. As $L{N^2}$ and ${L^2}tr\{ [{({S^{'}})^T}{S^{'}}]\} $ are const, equation (20.a) is turned into the expression in equation (20).
\[\_\_\_\_\_\_\_\_\_\_\_\_\_\_\_\_\_\_\_\_\_\_\_\_\_\_\_\_\_\_\_\_\_\_\_\_\_\_\_\_\_\_\_\_\_\_\_\_\_\_\_\_\_\_\_\_\_\_\_\_\_\_\_\_\_\_\_\_\_\_\_\_\_\_\_\_\_\_\_\_\_\_\_\_\_\_\_\_\_\_\_\_\_\_\_\_\_\_\_\_\_\_\_\]
\end{figure*}
Then, equation (20) is rewritten as follows
\begin{equation}
\begin{array}{l}
 \mathop {\max }\limits_{{W_{x'}},{W_{y'}}} tr\{ 2[({W_{x'}}^T{x'^T}{A^{1/2}}){S^{'}}({A^{1/2}}y'{W_{y'}})]\},  \\
 s.t. \\
 \begin{array}{*{20}{c}}
   {{{({A^{1/2}}x'{W_{x'}})}^T}{A^{1/2}}x'{W_{x'}} = N{I_L}},  \\
   {{{({A^{1/2}}y'{W_{y'}})}^T}{A^{1/2}}y'{W_{y'}} = N{I_L}}.  \\
\end{array} \\
 \end{array}
\end{equation}
With application of Lagrange multiplier to equation (21), it yields
\begin{equation}
\begin{array}{l}
 J({W_{x'}},{W_{y'}}) = tr\{ 2[({W_{x'}}^T{x'^T}{A^{1/2}}){S^{'}}({A^{1/2}}y'{W_{y'}})]\}  -  \\
 \frac{\lambda }{2}\{ [{({A^{1/2}}x'{W_{x'}})^T}{A^{1/2}}x'{W_{x'}}) + {({A^{1/2}}y'{W_{y'}})^T}{A^{1/2}}y'{W_{y'}}]\\
  - 2NI_L^{}\}.  \\
 \end{array}
\end{equation}
To find the maximum of (22), it satisfies the following relation
\begin{equation}
 \frac{{\partial J({W_{x'}},{W_{y'}})}}{{\partial {W_{x'}}}} = 0,
\end{equation}
and
\begin{equation}
 \frac{{\partial J({W_{x'}},{W_{y'}})}}{{\partial {W_{y'}}}} = 0.
 \end{equation}
Thus, the solution to equation (22) is converted to (25)
\begin{equation}
\left[ {\begin{array}{*{20}{c}}
   0 & {{R_{x'y'}}}  \\
   {{R_{y'x'}}} & 0  \\
\end{array}} \right]W = \lambda \left[ {\begin{array}{*{20}{c}}
   {{R_{x'x'}}} & 0  \\
   0 & {{R_{y'y'}}}  \\
\end{array}} \right]W,
 \end{equation}
where
\begin{equation}
{R_{x'x'}} = {x'^T}Ax',
\end{equation}
\begin{equation}
{R_{y'y'}} = {y'^T}Ay',
\end{equation}
\begin{equation}
{R_{x'y'}} = {x'^T}A^{1/2}S^{'}A^{1/2}y',
\end{equation}
\begin{equation}
{R_{y'x'}} = {R_{x'y'}}^T,
\end{equation}
\begin{equation}
W = \left[ \begin{array}{l}
 {W_{x'}} \\
 {W_{y'}} \\
 \end{array} \right].
\end{equation}
Then, equation (25) is addressed by the generalized eigenvalue (GEV) problem. Based on the aforementioned description, the proposed multi-modal hashing with discriminative canonical correlation maximization (MH-DCCM) method is summarized in \textbf{Algorithm 1}.
\begin{algorithm}[htb]
\caption{The MH-DCCM Method}
\label{alg:Framwork}
\begin{algorithmic}
\REQUIRE ~~\\
\textbf{*} Calculate the zero-mean sets $x'$ and $y'$ from the original data sets;\\
\textbf{*} Construct $A$ and $S^{'}$ with the semantic label information;\\
\ENSURE ~~\\
\STATE \textbf{*} Calculate the matrices $R_{x'x'}$, $R_{y'y'}$, $R_{x'y'}$ and $R_{y'x'}$.
\label{ code:fram:extract }
\STATE \textbf{*} Find the solutions to (25).
\RETURN $W_{x'}$ and $W_{y'}$.
\end{algorithmic}
\end{algorithm}
\subsubsection{Non-Canonical Correlation Maximization}
\begin{figure*}[hb]
\hrulefill
\begin{align*} \label{equ:BigWrited LiZi}
\begin{array}{l}
 \mathop {\min }\limits_{{w^t}_{x'},{w^t}_{y'}} {\left\| {({A^{1/2}}x'{{w^t}_{x'}}){{({A^{1/2}}y'{{w^t}_{y'}})}^T} - {R_t}} \right\|^2} \\
  = {min} (tr\{ [({A^{1/2}}x'{{w^t}_{x'}}){({A^{1/2}}y'{{w^t}_{y'}})^T} - {R_t}] {[({A^{1/2}}x'{{w^t}_{x'}}){({A^{1/2}}y'{{w^t}_{y'}})^T} - {R_t}]^T}\})  \\
 \end{array}
 \tag{34.a}
\end{align*}
Since the vectors ${w^t}_{x'}$ and ${w^t}_{y'}$ have been normalized to be const individually and $R_t$ is only associated with vectors ${w^1}_{x'},{w^2}_{x'},...,{w^{t - 1}}_{x'}$ and ${w^1}_{y'},{w^2}_{y'},...,{w^{t - 1}}_{y'}$, equation (34.a) is turned into the expression in equation (34).
\end{figure*}
Although we obtain a close-form solution to equation (25) by GEV, it results in the following practical problem. For the canonical correlation analysis, the `canonical' condition drives the solution to select projected dimensions with low variance, as shown in Figure 4. Nevertheless, it is acknowledged that there are obvious variances in different projected dimensions and the dimensions with larger-variance potentially contain more significant information [27, 77].
\begin{figure}[H]
\centering
\includegraphics[height=1.5in,width=3.4in]{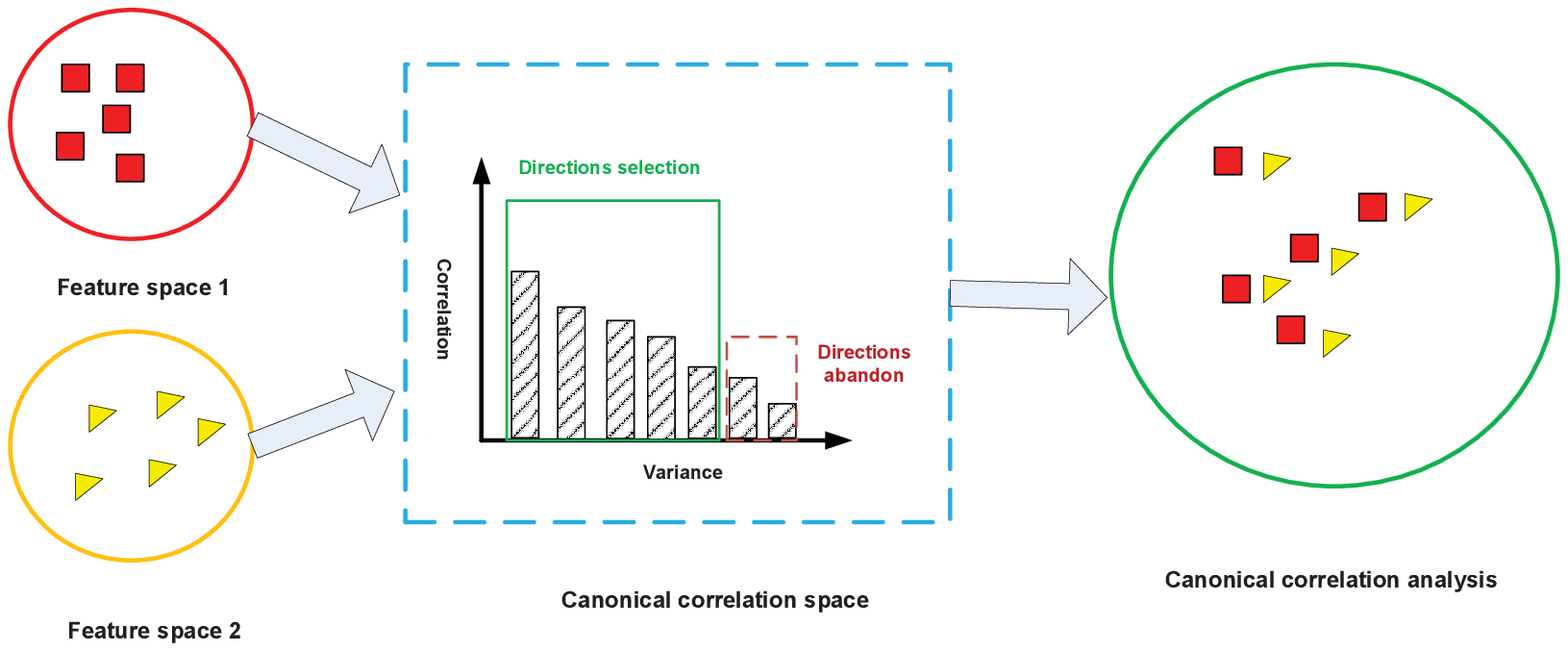}\\ Figure. 4 Directions selection by canonical correlation analysis\\
\end{figure}
Moreover, it is known that the non-canonical projection vectors are able to obtain better recognition accuracy than that of canonical ones [28]. As a result, a non-canonical solution is considered in this subsection. In this paper, to find the projected matrices ${W_{x'}}$ and ${W_{y'}}$ to minimize equation (16), an iterative algorithm is presented as follows:\\\indent Initially, suppose we have $t-1$ pairs of projected vectors ${w^1}_{x'},{w^2}_{x'},...,{w^{t - 1}}_{x'}$ and ${w^1}_{y'},{w^2}_{y'},...,{w^{t - 1}}_{y'}$. The next step is to obtain the $t$th vectors ${w^t}_{x'}$ and ${w^t}_{y'}$, which satisfy the normalized relation, $ {({A^{1/2}}x'{{w^t}_{x'}})^T}{A^{1/2}}x'{{w^t}_{x'}} = {({A^{1/2}}y'{{w^t}_{xy'}})^T}{A^{1/2}}y'{{w^t}_{y'}} = 1$.\\\indent Define a matrix ${R_t}$ in equation (31)
\begin{equation}
{R_t} = L{S^{'}}{\rm{ - }}\sum\limits_{k = 1}^{t - 1} {{\rm{sgn}}({A^{1/2}}x'{w^k}_{x'}){\rm{sgn}}{{({A^{1/2}}y'{w^k}_{y'})}^T}}.
\end{equation}
Then, equation (16) is converted into (32)
\begin{equation}
\mathop {\min }\limits_{{w^t}_{x'},{w^t}_{y'}} \left\| {{\rm{sgn}}({A^{1/2}}x'{w^t}_{x'}){\rm{sgn}}{{({A^{1/2}}y'{w^t}_{y'})}^T} - {R_t}} \right\|^2.
\end{equation}
Again, the method of spectral relaxation is performed on equation (32), leading to (33)
\begin{equation}
\mathop {\min }\limits_{{w^t}_{x'},{w^t}_{y'}} \left\| {({A^{1/2}}x'{w^t}_{x'}){{({A^{1/2}}y'{w^t}_{y'})}^T} - {R_t}} \right\|^2.
\end{equation}
Equation (33) is converted into (34)
 \begin{equation}
\begin{array}{l}
 \mathop {\min }\limits_{{w^t}_{x'},{w^t}_{y'}} {\left\| {({A^{1/2}}x'{{w^t}_{x'}}){{({A^{1/2}}y'{{w^t}_{y'}})}^T} - {R_t}} \right\|^2} \\
  = \mathop {\min }\limits_{{w^t}_{x'},{w^t}_{y'}} (- tr\{ 2[({{w^t}_{x'}}^T{x'^T}{A^{1/2}}){R_t}({A^{1/2}}y'{{w^t}_{y'}})]\} + const)  \\
 \end{array}
\end{equation}
The derivation of equation (34) is given in (34.a).\\
Equivalently, the solution to equation (34) is expressed in equation (35)
\begin{equation}
\begin{array}{l}
 \mathop {\max }\limits_{{{w^t}_{x'}},{{w^t}_{y'}}} tr\{ 2[({{w^t}_{x'}}^T{x'^T}{A^{1/2}}){R_t}({A^{1/2}}y'{{w^t}_{y'}})]\}. \\
 \end{array}
\end{equation}
Define the objective function of non-canonical correlation analysis ${D^{(t)}}_{x'y'}$ as follows
\begin{equation}
{D^{(t)}}_{x'y'} = {({A^{1/2}}x')^T}{R_t}({A^{1/2}}y').
\end{equation}
Then, equation (35) is rewritten as
\begin{equation}
\begin{array}{l}
 \mathop {\max }\limits_{{{w^t}_{x'}},{{w^t}_{y'}}} tr\{ 2[({{w^t}_{x'}}^T \cdot {D^{(t)}}_{x'y'} \cdot {{w^t}_{y'}})]\}. \\
 \end{array}
\end{equation}
In addition, since ${D^{(t)}}_{x'y'}$ satisfies the following relation
\begin{equation}
\begin{array}{l}
 {D^{(t)}}_{x'y'} = {({A^{1/2}}x')^T}{R_t}({A^{1/2}}y') \\
  = L{({A^{1/2}}x')^T}{S^{'}}({A^{1/2}}y') \\
  - \sum\limits_{k = 1}^{t - 1} {{{({A^{1/2}}x')}^T}{\rm{sgn}}({A^{1/2}}x'{w^k}_{x'}){\rm{sgn}}{{({A^{1/2}}y'{w^k}_{y'})}^T}} ({A^{1/2}}y') \\
  = {D^{(t - 1)}}_{x'y'} \\
  - {({A^{1/2}}x')^T}{\rm{sgn}}({A^{1/2}}x'{w^{t - 1}}_{x'}){\rm{sgn}}{({A^{1/2}}y'{w^{t - 1}}_{y'})^T}({A^{1/2}}y'), \\
 \end{array}
\end{equation}
and the previous vectors ${w^{t - 1}}_{x'}$ and ${w^{t - 1}}_{y'}$ are const for the $t$th iteration, the solutions (${w^t}_{x'}$ and ${w^t}_{y'}$) to equation (37) are also achieved by the GEV method.\\\indent
In this paper, set ${R_0} = S^{'}$ and the initial ${D^{(0)}}_{x'y'}$ is given as follows
\begin{equation}
\begin{array}{l}
 {D^{(0)}}_{x'y'} = {({A^{1/2}}x')^T}{R_0}({A^{1/2}}y') \\
  = {({A^{1/2}}x')^T}{S^{'}}({A^{1/2}}y') \\
  = {({A^{1/2}}x')^T}(2U{U^T} - {\bf{1}}\cdot{{\bf{1}}^T})({A^{1/2}}y') \\
  = 2[{({A^{1/2}}x')^T}U]{[{({A^{1/2}}y')^T}U]^T}\\
  -[{({A^{1/2}}x')^T}{\bf{1}}]{[{({A^{1/2}}y')^T}{\bf{1}}]^T}.\\
 \end{array}
\end{equation}
Thus, according to the aforementioned description, an iterative strategy is proposed and applied to the non-canonical correlation maximization method. Moreover, as both ${w^t}_x$ and ${w^t}_y$ are achieved by the GEV method, the proposed non-canonical method possesses a closed-form solution. Hence, it is not necessary to force hyper-parameters or stopping conditions on this non-canonical method. This is an important characteristic, especially in large scale problems.\\\indent Based on the above analysis and discussion, the proposed multi-modal hashing with discriminative non-canonical correlation maximization (MH-DNCCM) method is summarized in \textbf{Algorithm 2}.
\begin{algorithm}[htb]
\caption{The MH-DNCCM Method}
\label{alg:Framwork}
\begin{algorithmic}
\REQUIRE ~~\\
\textbf{*} Calculate the zero-mean sets $x'$ and $y'$ from the original data sets;\\
\textbf{*} Construct $A$ and $S^{'}$ with the semantic label information;\\
\ENSURE ~~\\
\STATE \textbf{*} Compute ${D^{(0)}}_{x'y'}$ according to $A$, $S^{'}$, $x'$ and $y'$;\\
\STATE \textbf{For t = 1 $\to$ Q ( Q = m + p) do}
\STATE \textbf{*} Calculate the generalized eigenvalue problem in equation (35).
\STATE \textbf{*} Obtain the eigenvector ${w^t}_{x'}$ and ${w^t}_{y'}$ associated with the largest eigenvalue.
\STATE \textbf{*} Construct ${D^{(t)}}_{x'y'}$ in equation (36).
\STATE \textbf{*} Substitute ${D^{(t)}}_{x'y'}$ to equation (37).
\STATE \textbf{End for}
\label{ code:fram:extract }
\RETURN ${W_{x'}} = [{w^1}_{x'},{w^2}_{x'},...{w^Q}_{x'}]$,\\
 \qquad \quad \    ${W_{y'}} = [{w^1}_{y'},{w^2}_{y'},...{w^Q}_{y'}]$.
\end{algorithmic}
\end{algorithm}
\section{Feature Extraction}
In this section, features extracted for evaluating the performance of the proposed MH-DCCM and MH-DNCCM are presented. Note, since multi-modal hashing is able to handle both multi-source and cross-modal feature representation, features generated from a single data modality but different algorithms are also utilized to verify the effectiveness of the proposed solutions. To demonstrate the generic nature of the feature representation methods, we conduct experiments based on features generated by different means, including classic features and DNN based features.
\subsection{Feature Extraction for Data Visualization}
In this work, we conducted experiments on the University of California Irvine (UCI) iris data set to visually compare the performance of the proposed solutions.
Specifically, a two-class sub-data set with species of `setosa' and `versicolor' by two features, `sepal length' and `sepal width'.\\
(a)The sepal length.\\
(b)The sepal width.
\subsection{Feature Extraction for Face Recognition}
In this paper, two kinds of classical features are used in fusion for face recognition [29-31]:\\
(c) The histogram of oriented gradients (HOG).\\
(d) Gabor features (6 orientations and 4 scales).
\subsection{Feature Extraction for Object Recognition}
Recently, a great number of DNN based algorithms have been proposed for object recognition. To further demonstrate the power of information fusion via the proposed strategy, a relatively simple DNN model, VGG--19 [75], is employed for DNN based feature extraction and compared with state-of-the-art DNN based algorithms. In this paper, two fully connected layers \textbf{fc7} and \textbf{fc8} of the VGG-19 model are utilized as DNN based features.\\
(e) The DNN feature from \textbf{fc7} layer.\\
(f) The DNN feature from \textbf{fc8} layer.
\subsection{Feature Extraction for Text-Image Recognition}
Text-image recognition has drawn notable attention within machine learning and data mining communities [33-35]. In this paper, we extract two kinds of features for text-image recognition: the bag-of-visual SIFT (BOV-SIFT) vector from images [36] and the deep learning-based feature Deep Latent Dirichlet Allocation (DLDA) from texts [37].\\
(g) The BOV-SIFT feature.\\
(h) The DLDA feature.
\subsection{Feature Extraction for Audio Emotion Recognition}
Audio is usually considered as one of the most natural and passive types of factors in emotion recognition. Therefore, two audio features, Prosodic and MFCC, are used in this study for human audio emotion recognition [78-79].\\
(i) The Prosodic feature [80].\\
(j) The MFCC feature [80].
\section{Experimental Results and Analysis}
To demonstrate the effectiveness of MH-DCCM and MH-DNCCM for multi-modal feature representation, we conduct data visualization on the UCI iris data set, face recognition on the ORL face database, object recognition on the Caltech 256 database, cross-modal recognition on the Wiki database and audio emotion recognition on the eNTERFACE (eNT) emotion database, respectively. Note that, during the experiments, the related multi-modal feature fusion experiments are conducted with the same features for fair comparison. The recognition accuracy, which is calculated as the ratio of the number of correctly classified samples over the total number of testing samples, is utilized to evaluate the performance of different algorithms. Moreover, in this work, we conducted experiments with different code length, and the optimal results are reported.\\
\subsection{The UCI Iris Data}
In this subsection, we construct a two-class sub-data set with species of `setosa' (50 samples) and `versicolor' (50 samples) by two features, `sepal length' and `sepal width'. Then, the distributions of the two features with the chosen 100 samples from the two species are plotted in Figure. 5. The horizontal coordinate is the index of samples and the vertical coordinate denotes the values of the two features.\\
\centerline {\includegraphics[height=1.42in,width=3.8in]{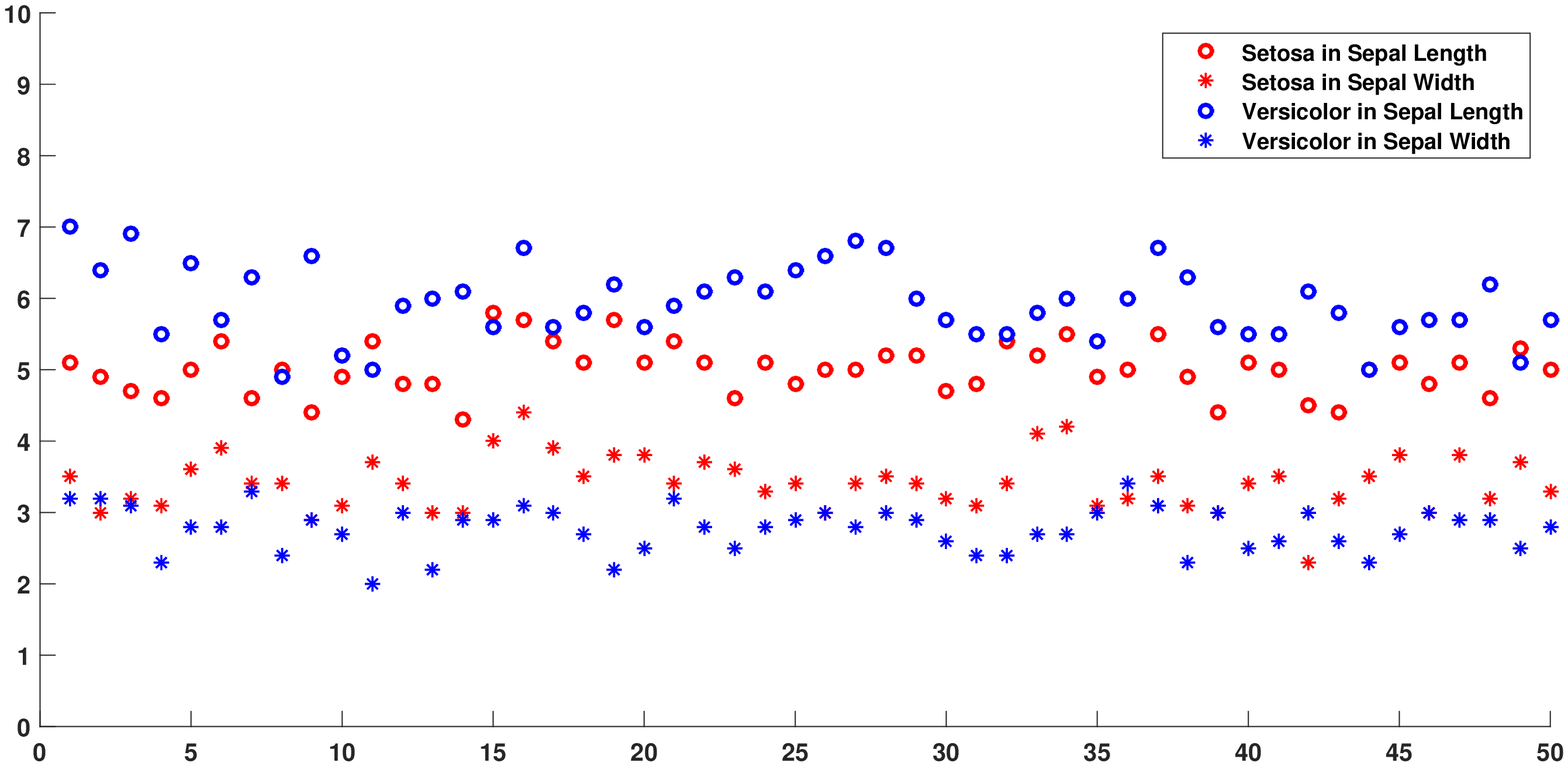}}\\ \centerline {Figure. 5. The original distribution of the two features}\\

Next, we conducted experiments by the semantic similarity, the discriminative correlation maximization (DCM), the proposed MH-DCCM and MH-DNCCM with the distributions plotted in Figure. 6 (a), (b), (c) and (d). Among them, the horizontal coordinate is the index of samples and the vertical coordinate stands for the projection of the original features in the semantic similarity space, DCM space, MH--DCCM space and MH--DNCCM space, respectively.\\
\centerline {\includegraphics[height=1.42in,width=3.8in]{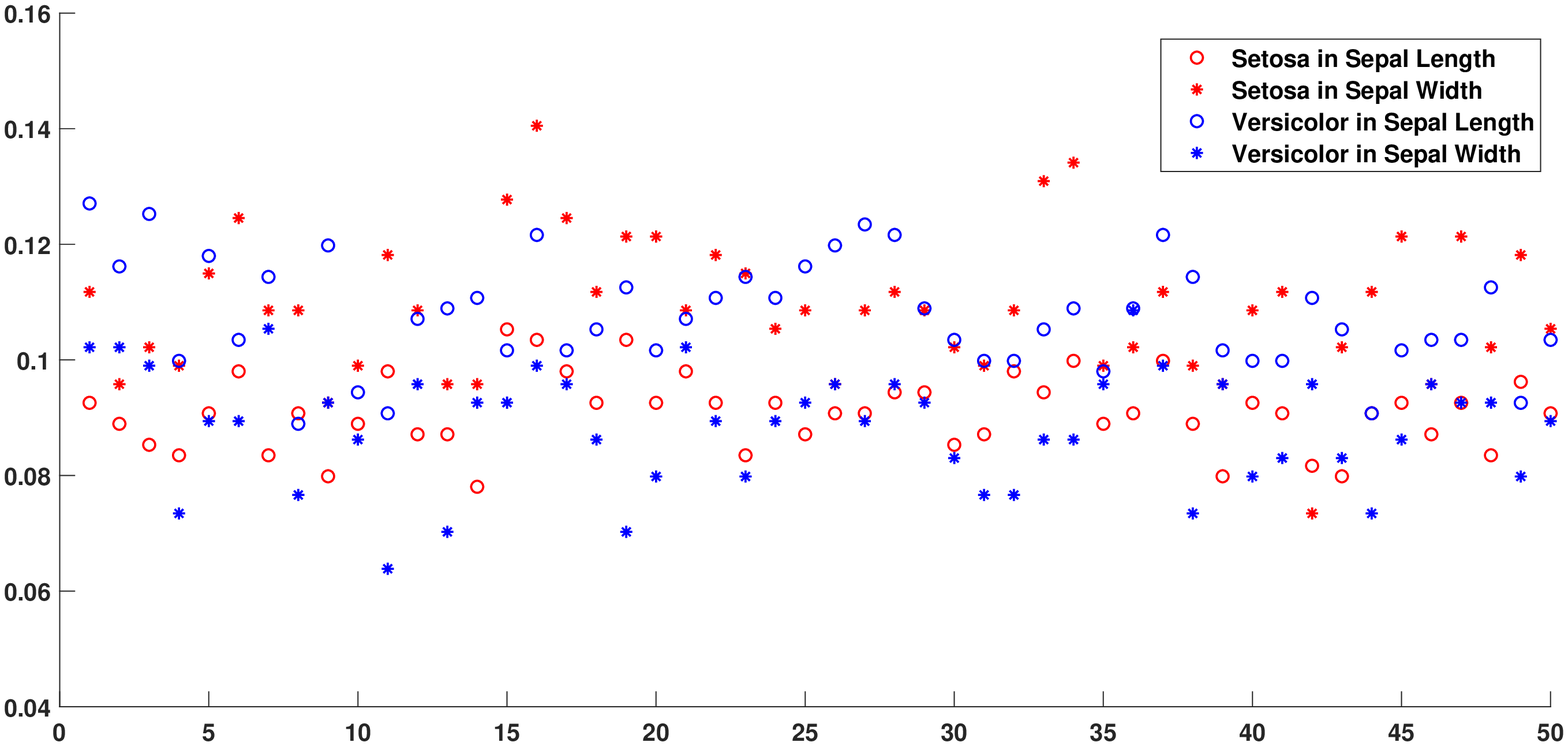}}\\ \centerline {Figure. 6(a). The visualization of semantic similarity}\\
\centerline {\includegraphics[height=1.42in,width=3.8in]{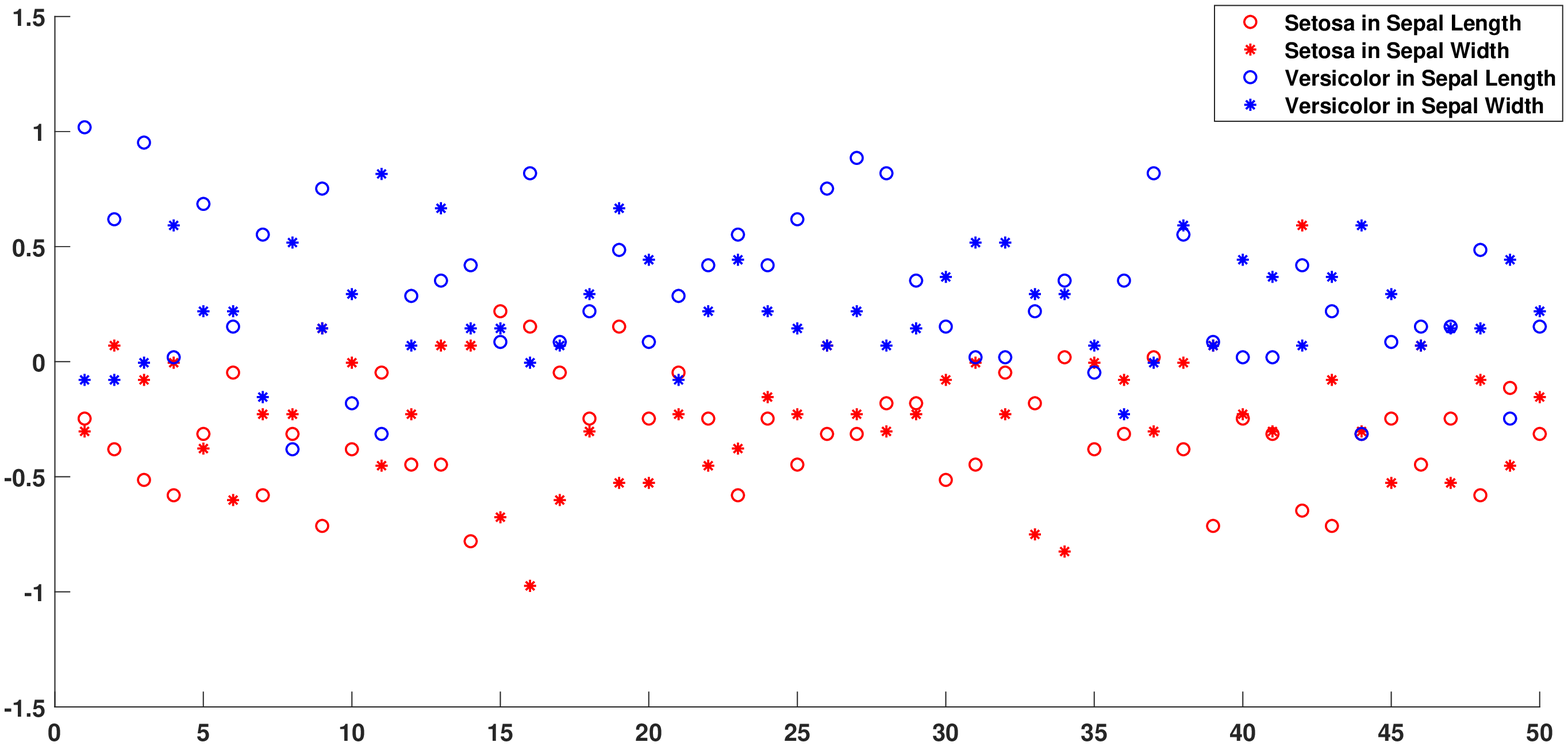}}\\ \centerline {Figure. 6(b). The visualization of DCM}\\
\centerline {\includegraphics[height=1.42in,width=3.8in]{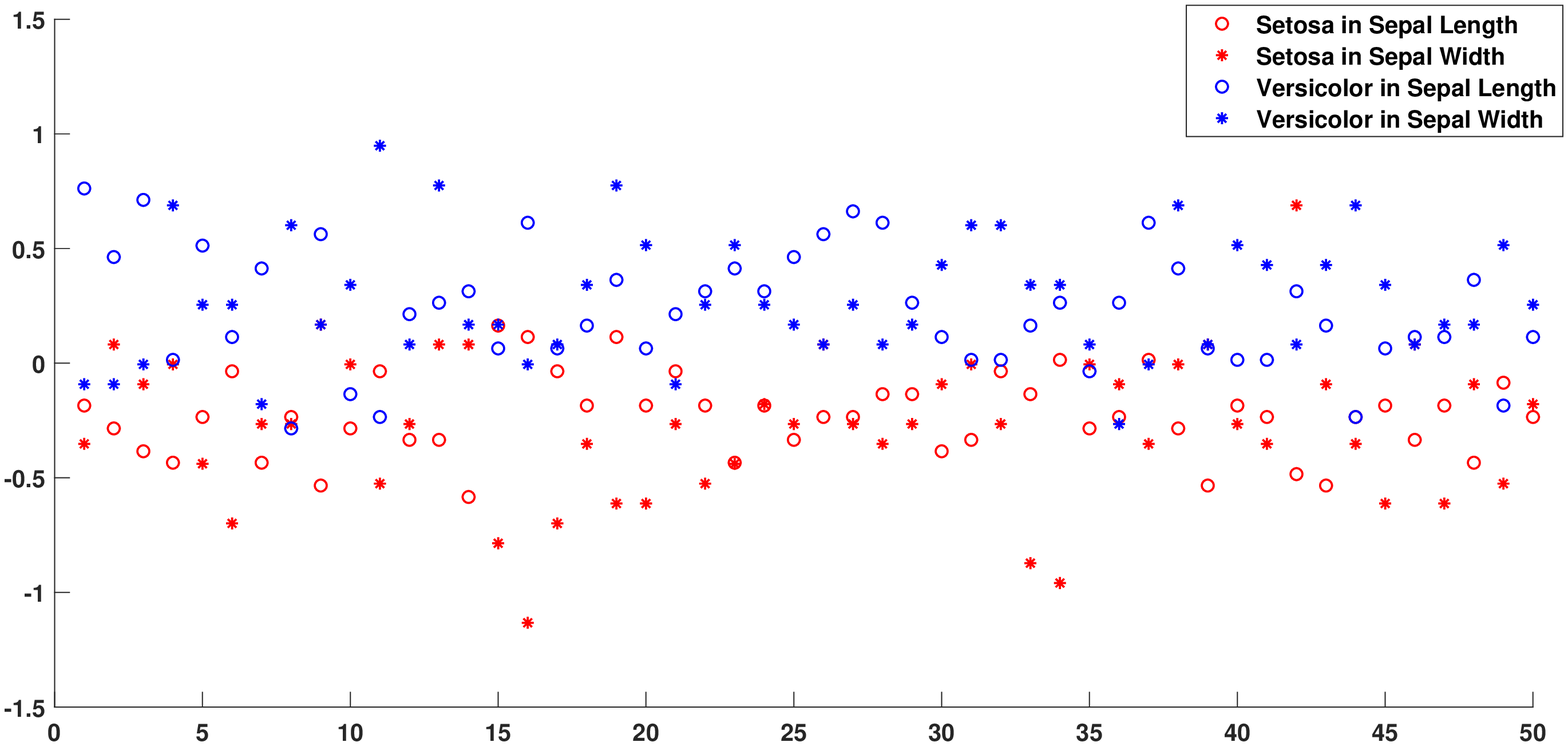}}\\ \centerline {Figure. 6(c). The visualization of MH-DCCM}\\
\centerline {\includegraphics[height=1.42in,width=3.8in]{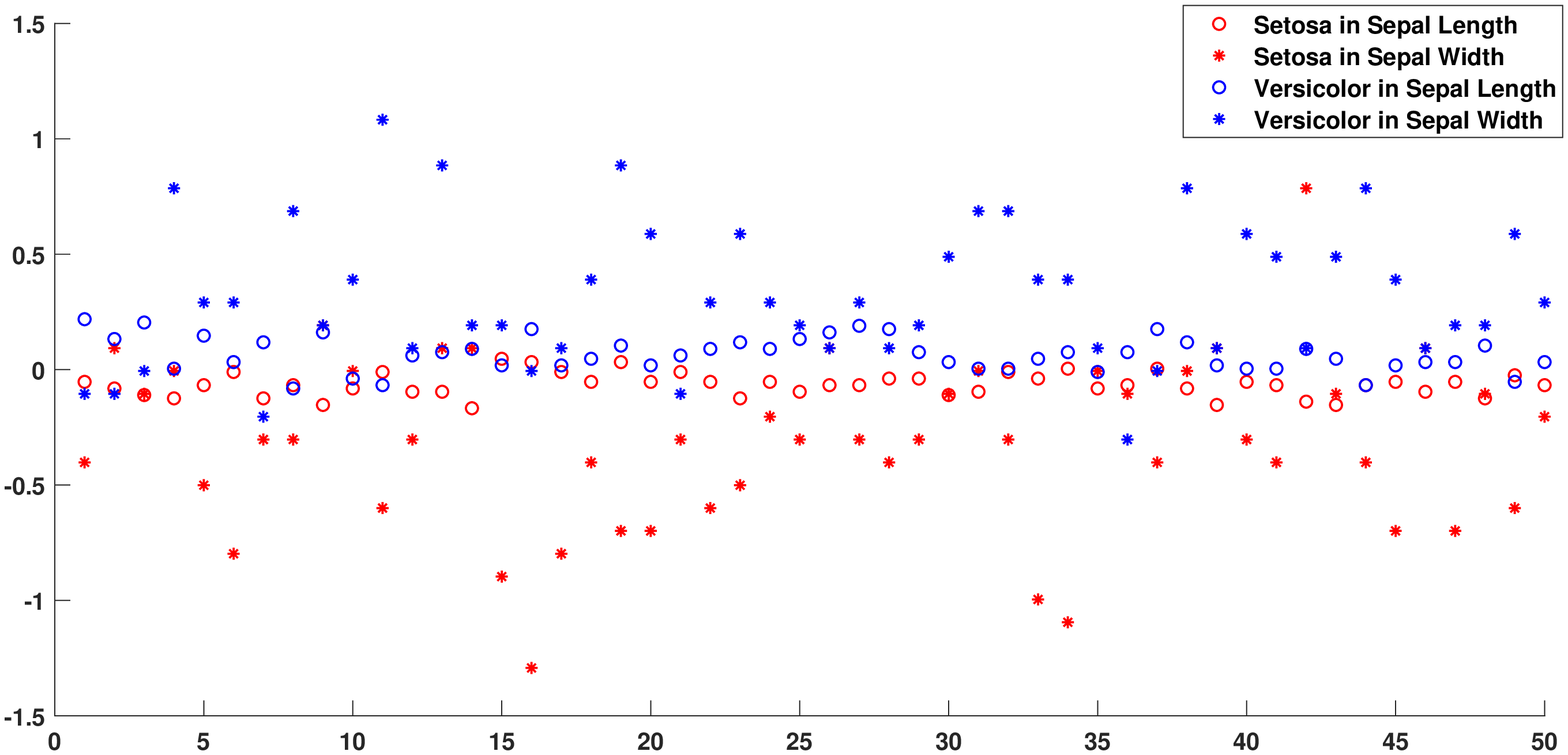}}\\ \centerline {Figure. 6(d). The visualization of MH-NDCCM}

Figure. 6 (a), (b), (c) and (d) show the visualization of the given two-class data sets by the semantic similarity, discriminative correlation maximization, MH-DCCM and MH-NDCCM. From these figures, it is observed that:\\
(1) All methods yield better feature representations than the original distribution.\\
(2) The semantic similarity criterion is able to minimize the distance between different features but ignores the discriminant representations.\\
(3) Benefiting from strengths of the semantic similarity and discriminative correlation maximization, MH-DCCM brings out more discriminant representations than semantic similarity alone. \\
(4) Compared with the MH-DCCM solution, MH-NDCCM explores more discriminant information.\\

In summary, by minimizing the semantic similarity among different features and exacting intrinsic discriminative representations across multiple data features with discriminative correlation maximization analysis jointly, MH-DCCM is able to generate more discriminant representations. According to the analysis in `Non-Canonical Correlation Maximization', since more significant information is potentially stored in non-canonical projections [28], the non-canonical projection vectors are capable of obtaining better recognition accuracy than that of canonical ones.
\subsection{The ORL Database}
The ORL consists of images from 40 persons, with each person providing 10 different samples. Each sample is normalized with size 64 $\times$ 64. In this database, some samples were obtained under different conditions, such as illumination, facial expression, and facial detail [99-102]. Some samples from the ORL database are given in Figure. 7. \\ \hspace*{\fill} \\
\centerline {\includegraphics[height=1.250in,width=2.8in]{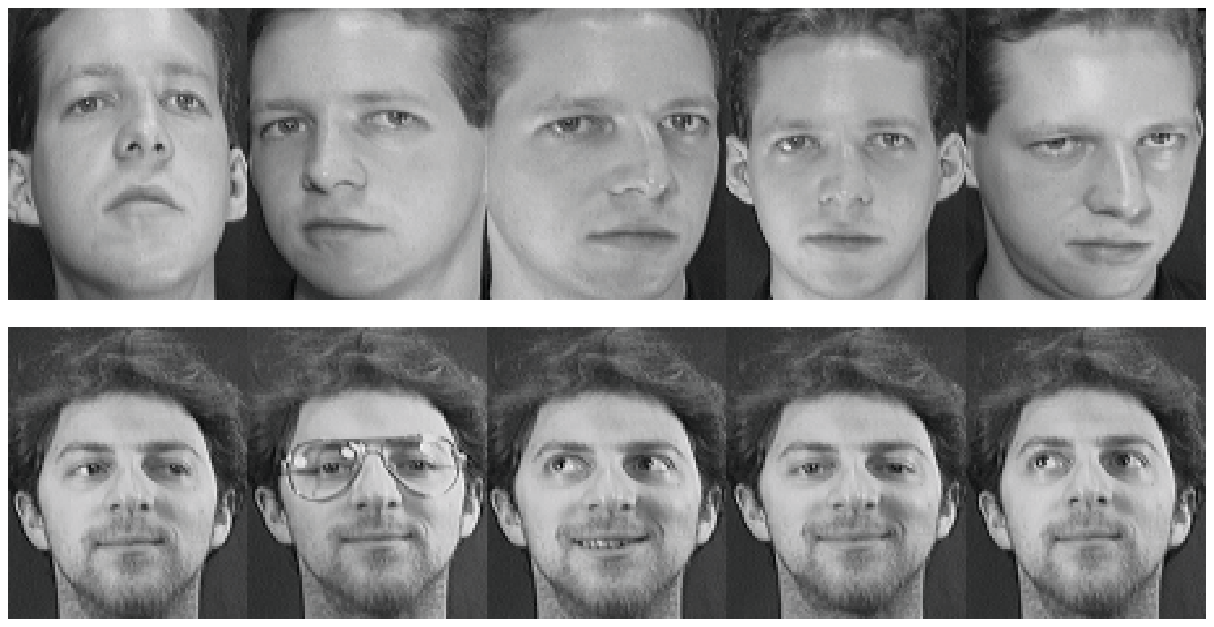}}\\ \centerline {Figure. 7. Samples from the ORL database}\\

During the experiments, we randomly selected 200 images (five from each person) to form the training subset while the remaining images were used for testing. The performance of HOG and Gabor features is individually evaluated first and results are shown in TABLE I. 
\begin{table}[h]
\normalsize
\renewcommand{\arraystretch}{1.0}
\caption{\normalsize{The performance with a single feature for face recognition}}
\setlength{\abovecaptionskip}{0pt}
\setlength{\belowcaptionskip}{10pt}
\centering
\tabcolsep 0.07in
\begin{tabular}{cc}
\hline
\hline
Feature & Accuracy\\
\hline
 HOG (ORL) &90.50\%\\
 Gabor (ORL) &88.50\%\\
\hline
\hline
\end{tabular}
\end{table}

Next, we conducted experiments with MH-DCCM \& MH-DNCCM methods and the results are given in Table II. For comparison, the performance of state-of-the-art algorithms [22, 25, 38-48, 70-72] is also presented in TABLE II. The comparison clearly shows the superiority of the proposed algorithms.
\vspace*{-10pt}
\begin{table}[h]
\normalsize
\renewcommand{\arraystretch}{1.0}
\caption{\normalsize{The performance with different algorithms on the ORL database}}
\setlength{\abovecaptionskip}{0pt}
\setlength{\belowcaptionskip}{10pt}
\centering
\tabcolsep 0.073in
\begin{tabular}{cc}
\hline
\hline
Methods  &  Accuracy \\
\hline
\textbf{MH-DNCCM}  &\textbf{98.00\%} \\
\textbf{MH-DCCM}  &\textbf{97.50\%} \\
 SCM-Seq [22] &95.50\% \\
 DSR [41] &94.50\% \\
 CRC [42] &88.50\% \\
 BULDP [70] &88.80\% \\
 PIM [71] &89.50\% \\
 L1LS [43] &92.50\% \\
 DALM [44] &90.00\% \\
 SOLDE-TR [45] &95.03\% \\
 GDLMPP [46]   &94.50\% \\
 CNN [47] &95.00\% \\
 PCANet [48] &96.50\% \\
 CS-SRC [72] &96.00\% \\
 DCCA [25]  &97.00\% \\
 LCCA [40] &95.50\% \\
 CCA [39] &94.50\% \\
 Serial Fusion [38] &77.50\% \\
\hline
\hline
\end{tabular}
\end{table}
\subsection{The Caltech 256 Database}
The Caltech 256 database contains 257 classes, including one background clutter class. Since Caltech 256 database represents a varying set of illumination, movements, backgrounds, etc., it is a challenging object recognition database [103-106]. The classes are hand-picked to represent a wide variety of natural and artificial objects in various settings. Several sample images from the Caltech 256 database are given in Figure. 8. In addition, for fair comparison, the experimental settings are similar to the other studies [49-63]. Specifically, 30 images are chosen from each class to construct the training subset. Then, the performance of two fully connected layers \textbf{fc7} and \textbf{fc8} is tabulated in TABLE III.
\begin{figure}[H]
\centering
\includegraphics[height=1.4in,width=3.0in]{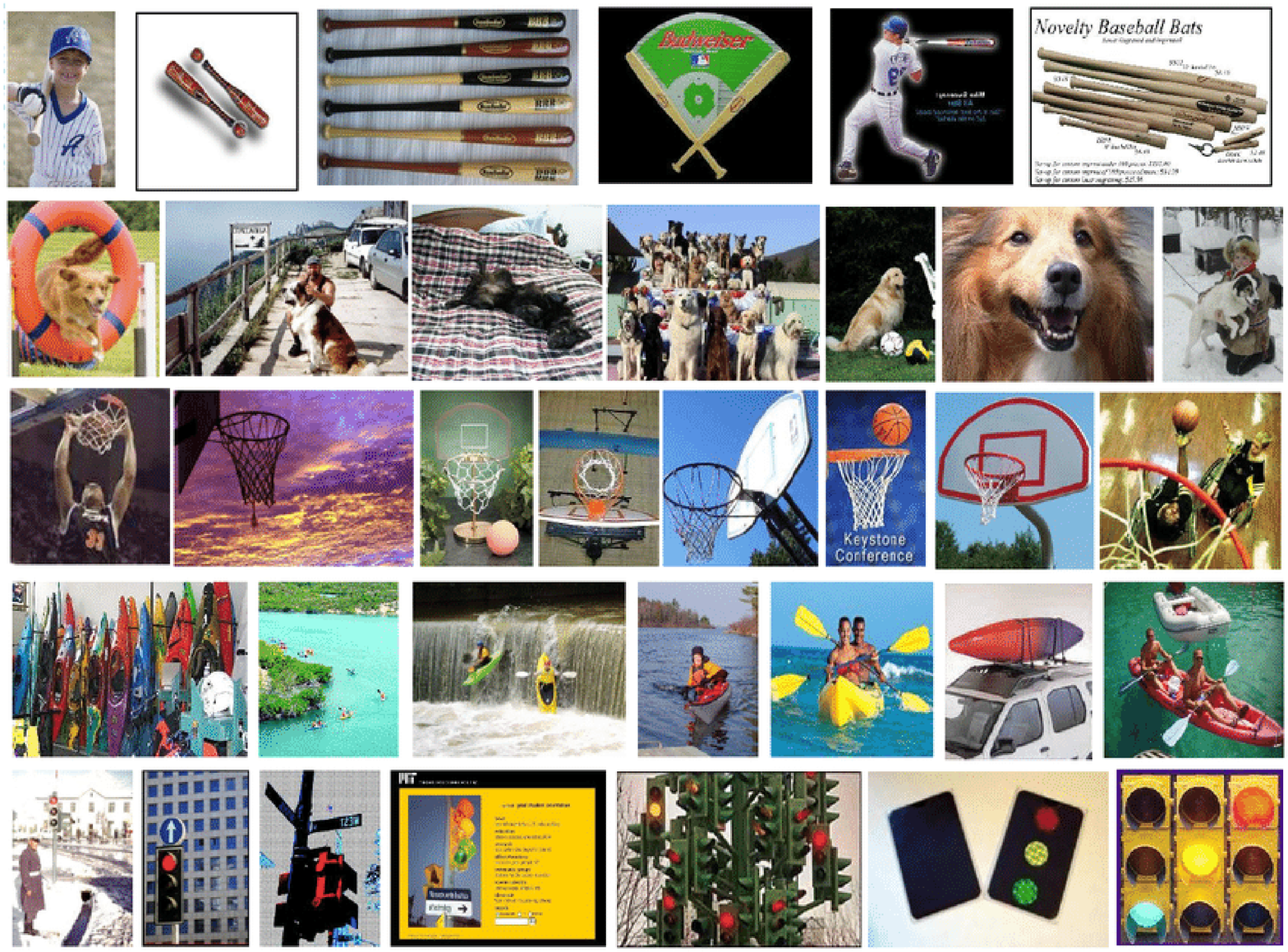}\\ Figure. 8 Some samples from the Caltech 256 database\\\indent
\end{figure}
\vspace*{-10pt}
\begin{table}[h]
\normalsize
\renewcommand{\arraystretch}{1.0}
\caption{\normalsize{The performance of \textnormal{fc7} and \textnormal{fc8} (Caltech 256)}}
\setlength{\abovecaptionskip}{0pt}
\setlength{\belowcaptionskip}{10pt}
\centering
\tabcolsep 0.07in
\begin{tabular}{ccc}
\hline
\hline
Feature & Accuracy \\
\hline
\textbf{fc7} & 59.81\% \\
\textbf{fc8} & 59.66\% \\
\hline
\hline
\end{tabular}
\end{table}

Then, \textbf{fc7} and \textbf{fc8} features are utilized to generate the new feature representation by the proposed MH-DCCM \& MH-DNCCM and the performance is shown in TABLE IV. To demonstrate the effectiveness of the proposed strategy, the performance by state-of-the-art methods are also reported in TABLE IV. Again, MH-DNCCM outperforms state-of-the-art methods in object recognition, including SML and DNN based algorithms, such as ResNet [62--63]. 
\vspace*{-10pt}
\begin{table}[h]
\normalsize
\renewcommand{\arraystretch}{1.00}
\caption{\normalsize{The Performance with state-of-the-art on the Caltech 256 database}}
\setlength{\abovecaptionskip}{0pt}
\setlength{\belowcaptionskip}{10pt}
\centering
\tabcolsep 0.07in
\begin{tabular}{ccc}
\hline
\hline
Methods & Accuracy \\
\hline
\textbf{MH-DNCCM}  &\textbf{81.03\%}\\
\textbf{MH-DCCM}  &\textbf{78.99\%}\\
HMML [49] &64.06\% \\
CMFA-SR [50] &71.44\% \\
DGFLP [51] &69.17\% \\
DCS [52] &69.86\% \\
SROSR [53] &75.60\% \\
BMDDL [54] &59.30\% \\
LLKc [55] &72.09\% \\
AL-ALL [56] &74.20\% \\
LMCCA [40] & 76.52\% \\
ISC-LG [57] &50.62\% \\
BLF-FV [58] &51.42\% \\
FSDH [96] &52.50\% \\
OCB-FV [59] &53.15\% \\
LLC-SVM [60] &70.70\% \\
SWSS-VGG [61] &73.56\% \\
Hybrid1365-VGG [87] &76.04\% \\
ResNet152 [62] &78.00\% \\
SSAH-Q [95] &78.60\% \\
ResFeats152 + PCA-SVM [63] &79.50\% \\
DMCCA [25] &80.32\% \\
\hline
\hline
\end{tabular}
\end{table}
\subsection{The Wiki Database}
The Wiki database is captured from various featured Wikipedia articles. In total, 2,866 documents are stored in image-text pairs and associated with supervised semantic labels from 10 classes [107-109]. In this paper, all documents are further divided into a training subset with 2173 documents and a testing subset with 693 documents, respectively. The performance of BOV-SIFT feature and DLDA feature is tested as a benchmark, as shown in TABLE V. Afterwards, BOV-SIFT and DLDA features are utilized for MH-DCCM \& MH-DNCCM, with their recognition rates reported in TABLE VI. The performance of state-of-the-art is also tabulated in TABLE VI. From TABLE VI, it is seen that the performance of the proposed methods is superior to those of state-of-the-art algorithms, SML and DNN, in text-image cross-modal recognition.
\begin{table}[h]
\normalsize
\renewcommand{\arraystretch}{1.0}
\caption{\normalsize{The performance of a single feature on the Wiki database}}
\setlength{\abovecaptionskip}{0pt}
\setlength{\belowcaptionskip}{10pt}
\centering
\tabcolsep 0.07in
\begin{tabular}{ccc}
\hline
\hline
Feature & Accuracy\\
\hline
BOV-SIFT & 17.46\% \\
DLDA & 66.23\% \\
\hline
\hline
\end{tabular}
\end{table}
\vspace*{-10pt}
\begin{table}[h]
\normalsize
\renewcommand{\arraystretch}{1.0}
\caption{\normalsize{The Performance with different algorithms on the Wiki database}}
\setlength{\abovecaptionskip}{0pt}
\setlength{\belowcaptionskip}{10pt}
\centering
\tabcolsep 0.07in
\begin{tabular}{ccc}
\hline
\hline
Methods & Accuracy \\
\hline
\textbf{MH-DNCCM}  &\textbf{69.26\%}\\
\textbf{MH-DCCM}  &\textbf{68.11\%}\\
DCML [73] & 55.40\% \\
SkeletonNet [68] & 56.31\% \\
SkeletonNet+pool [68] & 57.67\% \\
SSAH [92] & 59.40\% \\
CCA [39] & 62.77\% \\
LCCA [40] & 63.49\% \\
DCCA [25] & 64.79\% \\
SePH [93] & 62.59\% \\
SCM-Seq [22] & 63.92\% \\
MSC [64] & 61.48\% \\
MVLS [65] & 62.70\% \\
RE-DNN [66] & 63.95\% \\
DJSRH [94] & 65.83\% \\
$L_{2,1}$CCA [67] & 65.99\% \\
SPIB [69] & 66.50\% \\
\hline
\hline
\end{tabular}
\end{table}
\subsection{The eNT Database}
To further validate the effectiveness and verify the generic nature of the proposed methods for feature representation, we extend performance evaluation to  audio emotion recognition on the eNT database. This database includes video samples from 43 subjects, expressing six principal emotions [110-112]. The audio channel is recorded at a sampling rate of 48000 Hz. During the experiment, 456 audio samples from the eNT database are chosen. Then, the chosen 456 samples are further classified into the training subset and the testing subset, including 360 and 96 samples. Afterwards, the performance of Prosodic and MFCC features in audio channel is investigated and results are reported in TABLE VII.
\begin{table}[h]
\normalsize
\renewcommand{\arraystretch}{1.0}
\caption{\normalsize{The performance of a single feature on the {\upshape eNT} database}}
\setlength{\abovecaptionskip}{0pt}
\setlength{\belowcaptionskip}{10pt}
\centering
\tabcolsep 0.07in
\begin{tabular}{ccc}
\hline
\hline
Feature & Accuracy\\
\hline
Prosodic & 50.21\% \\
MFCC & 39.58\% \\
\hline
\hline
\end{tabular}
\end{table}\\\indent
Next, MH-DCCM \& MH-DNCCM are applied to Prosodic and MFCC features and experimental results are reported in TABLE VIII. The comparison with state-of-the-art [22, 25, 39-40, 81-86] is also presented in TABLE VIII. It is observed from the table that the proposed methods yield the best recognition accuracies.
\begin{table}[h]
\normalsize
\renewcommand{\arraystretch}{1.0}
\caption{\normalsize{The Performance with Different Methods on the {\upshape eNT} database}}
\setlength{\abovecaptionskip}{0pt}
\setlength{\belowcaptionskip}{10pt}
\centering
\tabcolsep 0.07in
\begin{tabular}{ccc}
\hline
\hline
Methods & Accuracy \\
\hline
\textbf{MH-DNCCM}  &\textbf{80.21\%}\\
\textbf{MH-DCCM}  &\textbf{78.13\%}\\
CCA [39]  & 63.54\%\\
FDA [81]  & 53.13\%\\
LCCA [40]  & 64.58\%\\
DCCA [25]  & 67.71\%\\
SCM-Seq [22]  & 71.88\%\\
DCNN [82]  & 72.80\%\\
DCNN-DTPM [82]  & 76.56\%\\
Alexnet (Fine-tuned) [83]  & 78.08\%\\
TSFFCNN [84] & 73.27\%\\
RF [85]  & 47.11\%\\
RBF-NN [86] & 75.89\%\\
\hline
\hline
\end{tabular}
\end{table}

According to aforementioned description, the experimental results clearly validated that the generated feature representation from multiple data sources is the winner in all the cases. Since the semantic similarity and intrinsic discriminant representations across different modalities are exploited jointly, it leads to improved performance compared with state-of-the-art. Moreover, since the non-canonical projection vectors potentially contain more important
information than the canonical ones, MH-DNCCM always achieves better performance than that of MH-DCCM on the evaluated data sets.
\section{Conclusions}
In this paper, a discriminative vectorial framework is proposed to generate a high quality feature representation from multi-modal features. The presented strategy is mathematically analyzed and further optimized according to canonical and non-canonical cases to implement feature coding for multi-modal feature representation. The experimental results clearly demonstrate that the proposed solutions outperform state-of-the-art on the data sets evaluated.

\end{document}